\documentclass[]{bytedance_seed}

\usepackage[toc,page,header]{appendix}

\usepackage{minitoc}

\usepackage{bbding}
\usepackage{multirow}
\usepackage{makecell}
\usepackage{titlesec}
\usepackage{pifont}
\usepackage{enumerate}
\usepackage[table]{xcolor}
\usepackage{xcolor}
\usepackage{color}
\usepackage{colortbl}
\usepackage{tablefootnote}
\usepackage{pifont}
\usepackage{makecell}
\usepackage{subcaption}
\usepackage{xspace}
\usepackage[most]{tcolorbox}
\newcommand{\ignore}[1]{}

\makeatletter
\DeclareRobustCommand\onedot{\futurelet\@let@token\@onedot}
\def\@onedot{\ifx\@let@token.\else.\null\fi\xspace}

\def\eg{\emph{e.g}\onedot}

\makeatother

\definecolor{takeaway}{RGB}{165, 209, 216}
\definecolor{takeawayTitle}{RGB}{57, 89, 163}

\title{Revisiting the Necessity of Lengthy Chain-of-Thought in\\ Vision-centric Reasoning Generalization}

\author[1,2,*]{Yifan Du}
\author[1,*]{Kun Zhou}
\author[1]{Yingqian Min}
\author[2]{Yue Ling}
\author[1,\dagger]{\\Wayne Xin Zhao}
\author[2,\dagger]{Youbin Wu}

\affiliation[1]{Renmin University of China}
\affiliation[2]{ByteDance Seed}

\contribution[*]{Equal Contribution}
\contribution[\dagger]{Corresponding authors}

\abstract{
We study how different Chain-of-Thought (CoT) designs affect the acquisition of the generalizable visual reasoning ability in vision-language models (VLMs). While CoT data, especially long or visual CoT such as ``think with image'', has been widely used to supervise intermediate reasoning, it remains unclear why specific CoT designs help and which ones truly support generalizable reasoning. \ignore{However, it is costly to construct or synthesize and may contain a complicated format that increases the risk of incorrect intermediate steps, hurting downstream reinforcement learning (RL).} To systematically evaluate this, we focus on a controlled maze-solving benchmark where reasoning rules are fully visual, difficulty can be tuned by grid size, and all the intermediate steps can be automatically generated. Using Qwen2.5-VL-7B under a standard SFT-then-RL pipeline, we compare three representative CoT formats: Language CoT, Grounding CoT (with spatial coordinate trajectories), and Visual CoT (with image manipulations). Our experiments reveal that visual and longer CoT mainly accelerate convergence but do not lift the final performance ceiling; concise CoT containing only essential grounding steps outperforms longer traces; and, strikingly, CoT retaining only the minimal grounding results generalizes best across different maze sizes. We further validate these insights on other vision-centric tasks. These findings highlight a ``short is long'' effect and provide practical guidance for constructing more generalizable SFT datasets for visual reasoning.
}

\date{\today}
\correspondence{Yifan Du at \email{yifandu1999@gmail.com}}

\checkdata[Project Page]{\url{https://github.com/RUCAIBox/Revisiting-Visual-CoT}}

\begin{document}
\maketitle


\section{Introduction}

Visual reasoning is emerging as a key capability for vision-language models (VLMs)~\cite{liu2023visual,li2024llava,wang2024qwen2,li2023evaluating}, enabling them to support broader real-world, vision-centric applications\footnote{\textbf{Vision-centric reasoning} refers to tasks where the core reasoning process must occur \emph{within the image}. The model must extract, track, and manipulate spatial or structural cues directly from the image.} rather than merely recognizing objects or generating captions~\cite{guo2024mammoth,team2025kimi,seed2025seed1_5vl}. To make such models truly reliable and versatile, it is essential to endow them with \emph{generalizable reasoning capabilities}. Instead of overfitting to a specific benchmark, generalizable reasoning aims to acquire abstract reasoning skills and reusable patterns that transfer across tasks, domains, and prompts, \eg, learning puzzle-solving rules on small mazes and simple sudoku games that can be transferred to larger and harder ones~\cite{xu2025visulogic,tong2025code2logic}.

To learn such generalizable visual reasoning capabilities, recent work widely employs Chain-of-Thought (CoT) data for supervised fine-tuning, to guide both the training and inference of VLMs~\cite{wang2025sota,deng2025openvlthinker}. CoT reasoning encourages the model to generate explicit intermediate steps instead of jumping directly to the answer, decomposing complex problems into a sequence of simpler sub-goals~\cite{wei2023cot}. By increasing the length of these CoT traces, with more detailed explanations, multi-step deliberation, and self-reflection, the VLM can reason more thoroughly and refine its predictions, often yielding more accurate and higher-quality results~\cite{du2025virgo,wang2025vl}. Moreover, supervised fine-tuning VLMs on long CoT–formatted data has been shown to further amplify these benefits, leading to significant performance improvements across diverse visual reasoning benchmarks~\cite{zhang2025thyme,wang2025simple,lai2025mini}. Among these methods, a special form of CoT expressed directly in the visual space, exemplified by o3’s ``think with image''~\cite{o3}, has proven particularly effective: this \emph{visual CoT} allows the model to manipulate the image itself (\eg, cropping regions or drawing marks to highlight evidence), making its reasoning process more aligned with how humans naturally think with images.

Despite these promising results, it remains unclear why these strategies help and which forms of CoT actually contribute to generalizable visual reasoning. In particular, different CoT paradigms externalize intermediate reasoning in fundamentally different ways (linguistic, spatial, and visual). It is unknown whether and how specific CoT designs contribute to learning generalizable visual reasoning ability. To close this gap, we conduct a rigorous and controlled comparison of three representative CoT formats: (1) \emph{Language CoT}, which follows the conventional LLM paradigm and expresses the entire reasoning process in natural language; (2) \emph{Grounding CoT}~\cite{wang2025traceable,shao2024visual}, where the model directly outputs a sequence of spatial coordinates so that the reasoning chain is implicitly encoded in the coordinate trajectory; and (3) \emph{Visual CoT}~\cite{o3,zheng2025deepeyes}, which manipulates the image via predefined operations (\eg, cropping or marking) and feeds the updated image back into the model to enable interleaved image–text reasoning. These formats reflect three different ways of externalizing intermediate reasoning (\eg, linguistic, spatial, and visual), allowing us to isolate how each affects learning and generalization.

To systematically and fairly compare these CoT strategies, we require a controlled task environment free from data contamination. Therefore, we select the maze, a classic vision-centric visual reasoning task, as our evaluation setting. The maze offers a clean and interpretable testbed: its underlying reasoning rule is fully expressed by the visual input, while the difficulty can be smoothly controlled by adjusting the grid size (\eg, from $4\times4$ to $6\times6$), and current VLMs still perform poorly on it (\eg, Qwen2.5-VL-7B~\cite{bai2025qwen25vltechnicalreport} achieves below 10\% success on $4\times4$ mazes), allowing us to focus on studying their visual reasoning ability rather than saturating performance. In addition, maze data are easy to synthesize, and both final solutions and intermediate steps (paths, coordinates, or annotated images) can be automatically generated, making it convenient to generate and filter large-scale CoT data. Built on this task, we design experiments that systematically test the widely used SFT-then-RL generalization paradigm under different CoT formats, by constructing CoT-specific cold-start datasets, performing Supervised Fine-Tuning (SFT) on Qwen2.5-VL-7B~\cite{bai2025qwen25vltechnicalreport} to obtain policy models for each format, and then further improving them with Reinforcement Learning (RL). After applying RL, our experiments reveal three key findings:

\begin{itemize}
\item \textbf{Visual and longer CoT accelerates but does not lift the ceiling.} Incorporating visual CoT and increasing CoT length can speed up convergence during training, yet the final performance plateau remains similar to (or only marginally better than) shorter-CoT baselines.
\item \textbf{Grounded short CoT surpasses verbose ones.} Short CoT containing essential grounding information achieves higher and more stable performance than longer, step-by-step CoT, suggesting that excessive intermediate explanation may not be useful for generalization.
\item \textbf{CoT with the least grounded results generalizes the best.} CoT formats that retain only the least amount of grounding results (\eg, sparse coordinate paths or final trajectories) achieve the strongest generalization across maze sizes.
\end{itemize}

In summary, our results show the ``short is long'' effect, suggesting that concise but well-grounded supervision better encourages reusable reasoning patterns. We also validate the effectiveness of short CoT on other visual reasoning tasks~(\eg, visual search and visual puzzle). It offers a practical guideline for constructing more generalizable SFT datasets for visual reasoning.

\section{Preliminary}
\begin{figure*}
    \centering
    \includegraphics[width=1\linewidth]{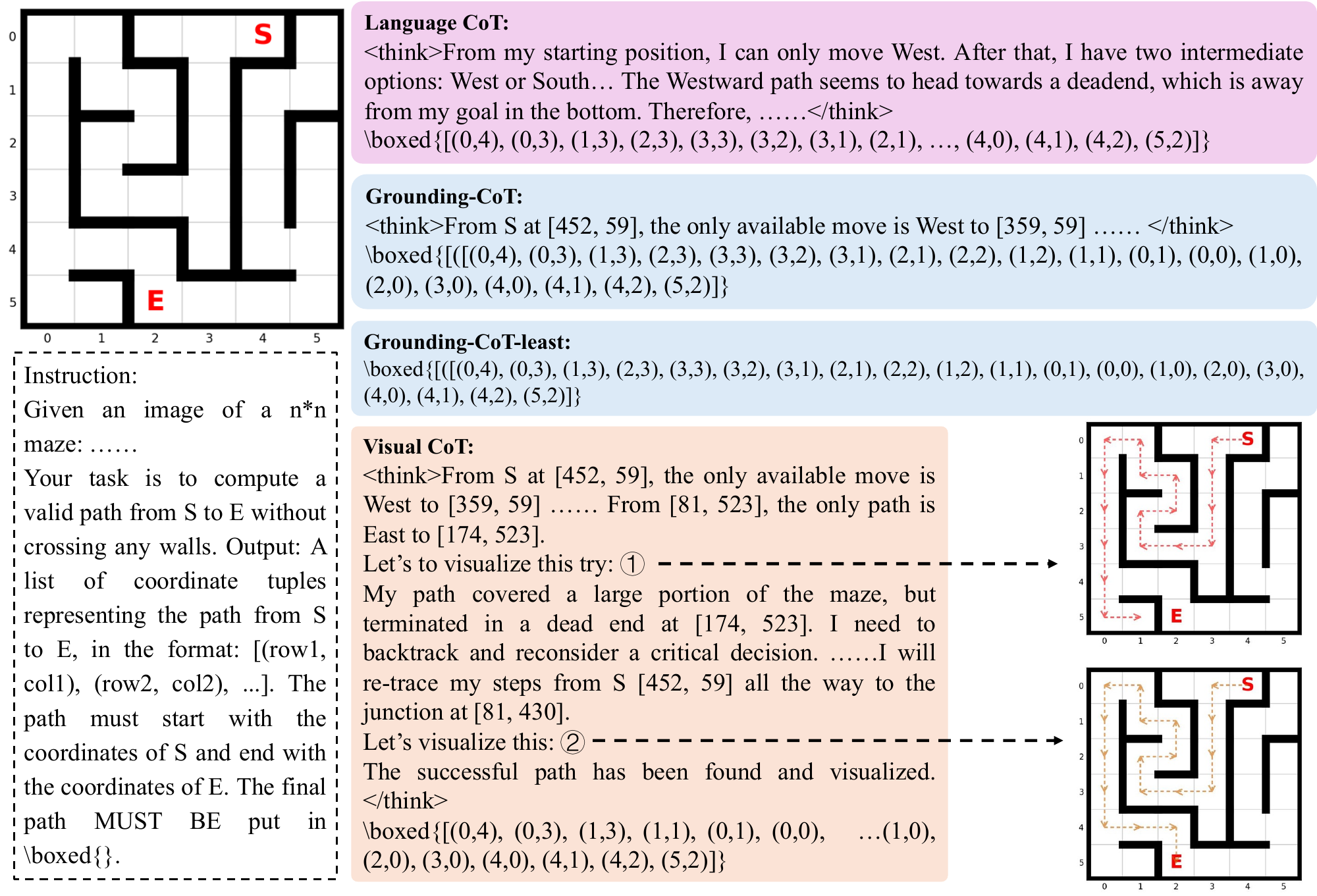}
    \caption{Illustration of four types of CoT reasoning strategies. We show examples of solving a 6$\times$6 maze navigation task.}
    \label{fig:maze}
\end{figure*}
We first formalize the Chain-of-Thought (CoT) reasoning process of a vision-language model (VLM). Given a textual query $Q$ and an image $I$ as input, the VLM $\pi_\theta$ generates a textual answer $A$. Its reasoning is decomposed into a sequence of intermediate steps (or thoughts) $R_T = {s_1, s_2, \ldots, s_T}$, where each $s_t$ denotes the reasoning state at time step $t$. The CoT strategy specifies how these intermediate reasoning states are represented.
In this paper, we mainly study the following three types of CoT strategies in visual reasoning tasks.

\subsection{Language CoT}
Language CoT follows the classical LLM paradigm, in which the reasoning process is expressed purely through textual tokens. Formally, the reasoning trajectory is $R_T^{\text{lang}} = { 
r_1^{(l)}, r_2^{(l)}, \ldots, r_T^{(l)}}, \quad r_t^{(l)} \in \mathcal{V}_{\text{text}}$, where $\mathcal{V}_{\text{text}}$ is the language vocabulary. At each reasoning step $t$, the model generates a reasoning step $r_t^{(l)} \sim \pi_\theta\left(\cdot | Q, I, R_{t-1}^{\text{lang}}\right)$ conditioned on the initial query $Q$, image $I$, and all the preceding reasoning steps $R_{t-1}^{\text{lang}}=r_1^{(l)}, r_2^{(l)}, \ldots,  r_{t-1}^{(l)}$. This formulation corresponds to explicit reasoning in the language space, without any manipulation on visual information. It has been widely used to enhance multimodal reasoning tasks that rely on stepwise logical deduction, such as visual math reasoning~\cite{wang2025vl,deng2025openvlthinker} and physical reasoning~\cite{chow2025physbench}.

\subsection{Grounding CoT}
Grounding CoT~\cite{wang2025traceable,shao2024visual} expresses the reasoning process by explicitly associating linguistic references with their corresponding visual evidence in the image\footnote{We name it as grounding CoT because it strictly localizes and grounds language concepts with the visual information at each step using a special format. While commonly-used language CoTs may include grounding steps, these are often sporadic and poorly structured.}. 
While bounding boxes are a popular grounding form in existing work~\cite{fan2025grit,sarch2025grounded}, we explore more flexible alternatives, including points, lines, and regions, which adapt to the reasoning context.
 Formally, each grounded element is represented by a spatial descriptor $g_k = (G_k, C_k)$, where $G_k\in\{\text{point, line, region}\}$ denotes the grounding type, and $C_k \subset [0,1]^2$ represents its spatial coordinates in the image:
\begin{equation}
C_k = 
\begin{cases} 
(x, y), &  G_k=\text{point} \\
\{(x_1, y_1), (x_2, y_2)\}, & G_k=\text{line} \\
\{(x_i, y_i)\}_{i=1}^n, & G_k=\text{region}
\end{cases}
\end{equation}
The trajectory using grounding CoT is denoted as $R_T^{\text{grd}} =  \left( r_1^{(g)}, g_1 \right), \left( r_2^{(g)}, g_2 \right), \dots, \left( r_T^{(g)}, g_T \right), \quad r_t^{(g)} \in \mathcal{V}_{\text{text}}$. At each reasoning step $t$, the model generates a reasoning step as $\left( r_t^{(g)}, g_t \right) \sim \pi_\theta\left(\cdot | Q, I, R_{t-1}^{\text{grd}}\right)$. By incorporating explicit yet flexible grounding information, the model learns to connect each reasoning step with the precise spatial evidence it refers to. This strategy enhances reasoning on vision-centric tasks such as counting~\cite{roberts2025zerobench}, spatial relation inference~\cite{wu2025reinforcing}, and path tracing~\cite{feng2025can}, where visual references may be points, lines, or regions.

\subsection{Visual CoT}
Visual CoT extends grounding CoT by enabling active visual manipulation during reasoning, with the help of specific tools or functions~\cite{zheng2025deepeyes,zhang2025thyme}. Instead of merely referencing visual elements, the model can dynamically operate on grounded visual evidence, such as highlighting a region, tracing a path, or cropping a subimage, and feed the modified visual context back into subsequent reasoning steps. Formally, at each reasoning step $t$, the model selects a grounding target $g_t$, and then an operation function $\phi_t(\cdot)$ acts on the current image $I_t$ conditioned on $g_t$:
\begin{equation}
    I_{t+1}=\phi_t(I_t, g_t),
\end{equation}
where $\phi_t$ may perform operations such as point marking, line drawing, and region cropping. The reasoning trajectory interleaves textual reasoning and visual updates: $R_T^{\text{vis}}= \left( r_1^{(v)}, g_1, I_1 \right), \left( r_2^{(v)}, g_2, I_2 \right), \ldots, \left( r_T^{(v)}, g_T, I_T \right) $. At each reasoning step $t$, the model generates an intermediate reasoning result $\left( r_t^{(v)}, g_t, I_t \right) \sim \pi_\theta\left(\cdot|Q,I,R_{t-1}^{\text{vis}}\right)$. This formulation enables iterative, interleaved image–text reasoning, where the model reasons about visual content, performs grounded visual actions, and refines its understanding through updated visual contexts.

\section{Experimental Setup}
\label{sec:exp_set}

In this section, we design our experiment to study \emph{how CoT strategies affect the learning of generalizable reasoning ability}.
To investigate it, we adopt the \emph{maze navigation task} as our evaluation testbed. This choice offers several advantages: (1) \textbf{Minimal interference from pretrained priors}: existing VLMs perform poorly on maze-related tasks, reducing confounding effects of strong pretrained capabilities; (2) \textbf{Pure visual reasoning}: maze solving mainly requires spatial reasoning and does not rely on external domain knowledge; and (3) \textbf{Controlled difficulty}: task complexity can be precisely tuned by adjusting maze size or path length. In this section, we introduce the experimental setup, including dataset construction, training strategy, and implementation details.


\subsection{Maze Data Construction}
\label{sec:data}
The maze is represented as an $N\times N$ grid $\mathcal{M} = \{(i, j) \mid i, j \in \{1, \ldots, N\}\}$, where all grids are reachable. Walls are defined between adjacent cells, rather than occupying cells themselves. Let $\mathcal{W} = \left\{ w_{(i,j) \to (i',j')} \mid (i,j), (i',j') \in \mathcal{M} \right\}$ denote the set of walls, where $w_{(i,j) \to (i',j')}=1$ indicates a wall between cells $(i,j)$ and $(i',j')$, and 0 otherwise. Given a start cell $s=(i_s,j_s)$ and an end cell $e=(i_e, j_e)$, the model can produce a stepwise reasoning trajectory $R_T = s_1, s_2, \ldots, s_T$ culminating in the correct path sequence $P=[(i_s, j_s), (i_2, j_2), \dots, (i_e, j_e)]$, satisfying $w_{(i_k,j_k) \to (i_{k+1},j_{k+1})}=0$. We visualize $\mathcal{M}$ as an image $I$, accompanied by an instruction $Q$, and the model is required to generate the reasoning process and the final path. Since current VLMs do not perform well in maze navigation and cannot even generate a reasonable thought process for maze navigation tasks~\cite{wu2025reinforcing}, we follow existing SFT-then-RL paradigm and first synthesize three types of data for supervised fine-tuning.

\paragraph{Language CoT Synthesis.} Language CoT in the maze navigation task describes the path through directions like ``north'', ``south'', ``west'', and ``east''. We employ language CoT because maze navigation inherently involves sequential spatial reasoning that can be naturally expressed as a series of linguistic direction tokens, allowing the model to align spatial movement with stepwise textual reasoning. To synthesize reasoning trajectories with language CoT, we first utilize a rule-based function to convert the path $P=[(i_s, j_s), (i_2, j_2), \dots, (i_e, j_e)]$ to a list of directions describing every step going from $s$ to $e$. Then we prompt Gemini-2.5-Pro~\cite{gemini2.5-pro} to synthesize the reasoning trajectories $R_T^{\text{lang}} =  r_1^{(l)}, r_2^{(l)}, \ldots, r_T^{(l)} $. We name these data as \textbf{L-CoT}. The detailed prompt is shown in Appendix, and an example of language CoT is shown in Figure~\ref{fig:maze}.

\paragraph{Grounding CoT Synthesis.} Grounding CoT in the maze navigation task refers to every stepped grid with its central coordinate.
To synthesize reasonable trajectories with grounding CoT, we first utilize a rule-based function to convert each grid $(i_k, j_k)$ in the path $P=[(i_s, j_s), (i_2, j_2), \dots, (i_e, j_e)]$ to its absolute coordinate in the image $[x_k, y_k]$. To enhance reasoning depth, we additionally introduce reflection patterns by synthesizing several incorrect paths (e.g., hitting a wall or entering a dead end) along with corresponding correction reasoning. Then we prompt Gemini-2.5-Pro~\cite{gemini2.5-pro} to synthesize the reasoning trajectories $R_T^{\text{grd}} =  \left( r_1^{(g)}, g_1 \right), \left( r_2^{(g)}, g_2 \right), \dots, \left( r_T^{(g)}, g_T \right) $, where $g_k=(\text{point}, [x_k, y_k])$. The detailed prompt is shown in Appendix. We name this dataset as \textbf{G-CoT}. It is worth noting that in the maze navigation task, the target output is a path from the start $s$ to the end $e$. It is equivalent to a grounding CoT, as it represents the sequence of visited grid positions without additional textual explanation or absolute coordinates. We therefore refer to this naturally aligned variant as \textbf{G-CoT-least}, indicating that the reasoning process is inherently embedded in the path sequence rather than being explicitly expressed. An example of the synthesized grounding CoT is shown in Figure~\ref{fig:maze}. 


\paragraph{Visual CoT Synthesis.} Visual CoT in the maze navigation task utilizes coordinates to refer to the stepped grid, and employ a predefined function to draw a line connecting these coordinates on the image. Compared to language CoT, visual CoT further enhances spatial reasoning by enabling explicit visual interaction, allowing the model to externalize its intermediate thoughts on the image and reason over the updated visual context. To synthesize reasoning trajectories with visual CoT, we first utilize a rule-based function to convert each grid $(i_k, j_k)$ in the path $P=[(i_s, j_s), (i_2, j_2), \dots, (i_e, j_e)]$ to its absolute coordinate in the image $[x_k, y_k]$. Then we define the operation function $\phi_t(\cdot)$ as a line-drawing operation on the image to reflect the drawing process to generate a sequence of intermediate images $\{I_1, I_2, \dots, I_T\}$, where each $I_t$ visualizes the partial path from the starting point $s$ to the current step $t$. Finally, we prompt Gemini-2.5-Pro~\cite{gemini2.5-pro} to synthesize the multimodal reasoning trajectories 
$R_T^{\text{vis}} =  \left( r_1^{(v)}, g_1, I_1 \right), \left( r_2^{(v)}, g_2, I_2 \right), \dots, \left( r_T^{(v)}, g_T, I_T \right) $ 
based on these interleaved image-text pairs. We name these data as \textbf{V-CoT}. The detailed prompt is shown in Appendix, and an example of visual CoT is shown in Figure~\ref{fig:maze}.


\subsection{Training Strategy}
To investigate how different CoT strategies affect model performance, we first perform supervised fine-tuning (SFT) using the three types of synthesized CoT data introduced in Section~\ref{sec:data}, obtaining three policy models with distinct reasoning styles. We then further train these models using reinforcement learning (RL) under the RLVR framework.

\paragraph{Supervised Fine-Tuning.}
We format the CoT data for SFT by enclosing the synthesized reasoning process within \texttt{<think></think>} tags and the final answer within \texttt{\textbackslash boxed\{\}} tags. Since the answers in G-CoT-least inherently contain the reasoning process, we do not separate them. For each CoT type, we synthesize 8K reasoning trajectories. In the visual CoT setting, the reasoning process is represented as interleaved image-text data, and the cross-entropy loss is computed only over textual tokens.

\paragraph{Reinforcement Learning.}
Although the fine-tuned VLM acquires the corresponding reasoning strategy, it still struggles to generalize to both in-domain and out-of-domain test data. To further enhance its reasoning robustness, we synthesize an additional 20K maze samples and apply reinforcement learning. The training dynamics reveal how different CoT strategies influence the optimization process. We employ Group Relative Policy Optimization (GRPO)~\cite{shao2024deepseekmath} to update the policy model. The reward function is defined as:
\begin{equation}
    r = \alpha \cdot r_{\text{acc}} + (1-\alpha)\cdot r_{\text{format}}
\end{equation}
The $r_{\text{acc}}$ is calculated based on whether the predicted path connects the start and end points without crossing walls, where we implement a rule-based function based on the maze structure. The $r_{\text{format}}$ is used to encourage the model to follow a specific format: enclosing the reasoning process in \texttt{<think></think>} and wrapping the final path using the symbol \texttt{\textbackslash boxed\{\}}. We set $\alpha$ to $0.1$ in our experiments.

\subsection{Implementation Details}
We adopt Qwen2.5-VL-7B~\cite{bai2025qwen25vltechnicalreport} as the base model. During the SFT stage, the model is fine-tuned on the synthesized CoT data using the LLaMAFactory~\cite{zheng2024llamafactory} framework. SFT is conducted for three epochs with a learning rate of $1\times10^{-5}$, a warm-up ratio of 0.1, and a batch size of 64. In the RL stage, we further optimize the model on the original maze dataset using the verl~\cite{sheng2024hybridflow} framework. The rollout batch size is set to 128, the mini-batch size to 32, and the number of rollouts to 8. RL training is continued until the model’s performance converges. Prior works in visual reinforcement learning~\cite{deng2025openvlthinker,leng2025mmr1,liu2025noisyrollout} typically train models for only a few hundred—or even just tens of—steps, which often leaves the model under-trained and makes its true performance ceiling unclear. To address this limitation, we train for up to 1000 RL steps, ensuring that the model fully converges and enabling a fair and reliable comparison across different CoT strategies. Throughout all training stages, we freeze the vision encoder and only update the parameters of LLM.
\section{Experimental Analysis}

\subsection{Experimental Framework}
To systematically investigate the mechanisms, benefits, and limitations of different CoT strategies in VLMs, our analysis is structured around the following research questions:

\begin{itemize}
    \item \textit{Q1: What benefits do these CoT strategies bring to VLMs in an unseen task?}

    \item \textit{Q2: What fundamental capability enables CoT to work in vision-centric tasks?}

    \item \textit{Q3: How is the generalizability of these CoT strategies?}
\end{itemize}

We first conduct experiments on the controlled maze navigation task and answer these questions in Section~\ref{sec:result}. We then extend our analysis to more realistic vision-centric tasks in Section~\ref{sec:real} to verify whether the observed mechanisms and benefits generalize beyond the maze setting.

\subsection{Experimental Results}

\label{sec:result}
\subsubsection{Visual CoT Improves Efficiency, Not Efficacy}
\label{sec:exp1}
\begin{figure*}
    \centering
    \begin{subfigure}{0.32\linewidth}
    \includegraphics[width=1\linewidth]{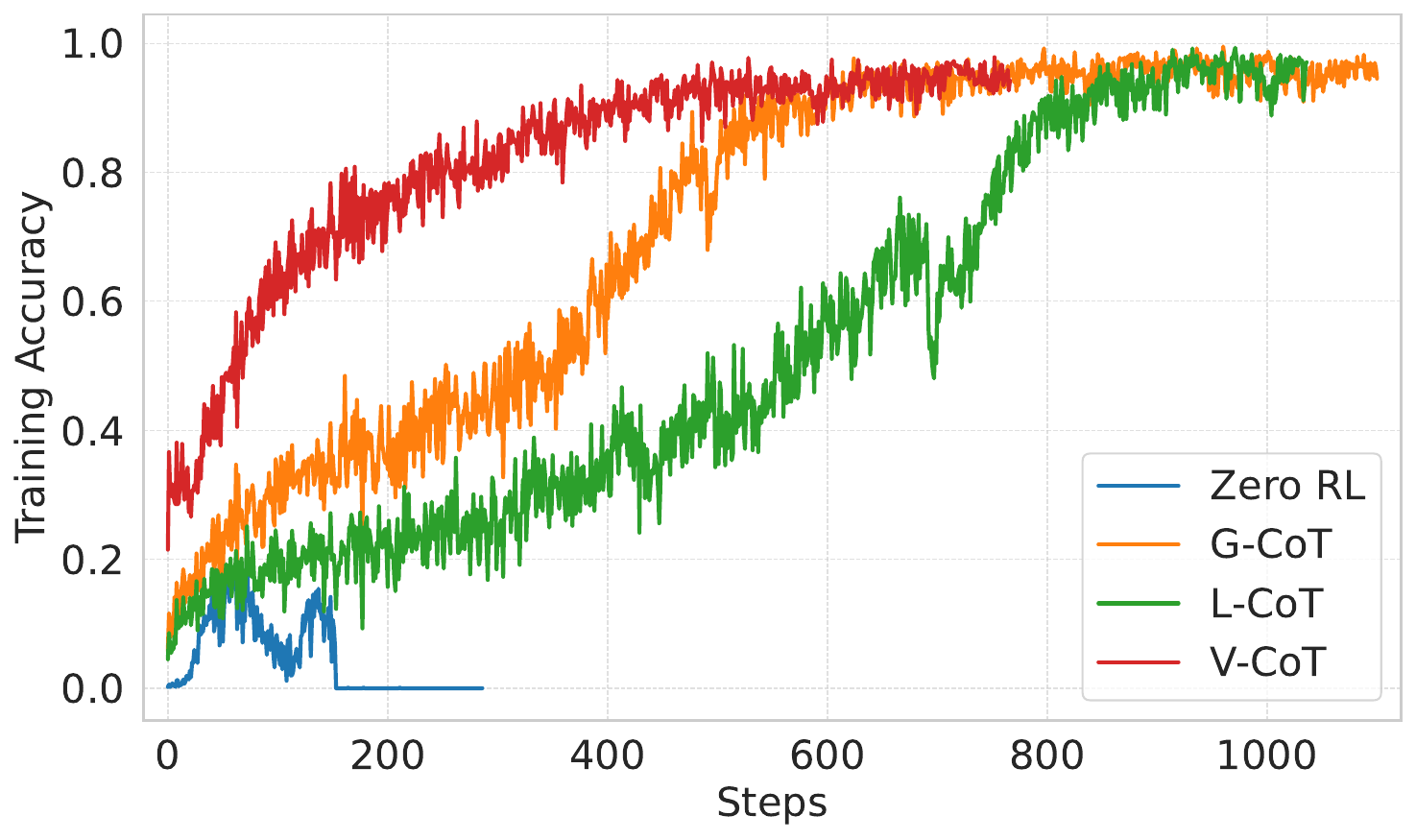}
    \caption{Training Accuracy on Mazes Sized $4\times4$ to $6\times6$.}
    
    \end{subfigure}
    \begin{subfigure}{0.32\linewidth}
    \centering
    \includegraphics[width=1\linewidth]{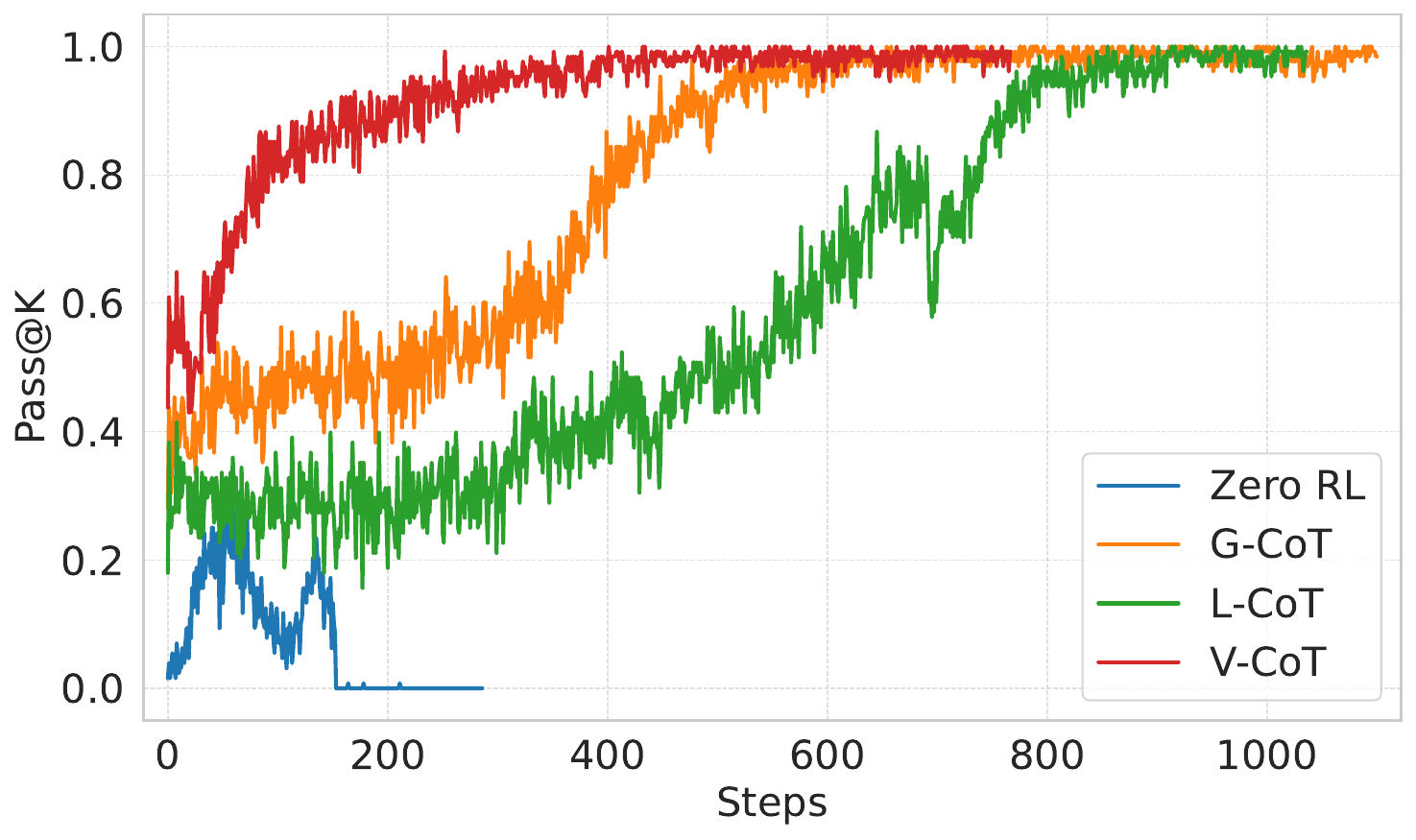}
    \caption{Pass@8 Training Accuracy on Mazes Sized $4\times4$ to $6\times6$.}
    
    \end{subfigure}
    \begin{subfigure}{0.32\linewidth}
    \centering
    \includegraphics[width=1\linewidth]{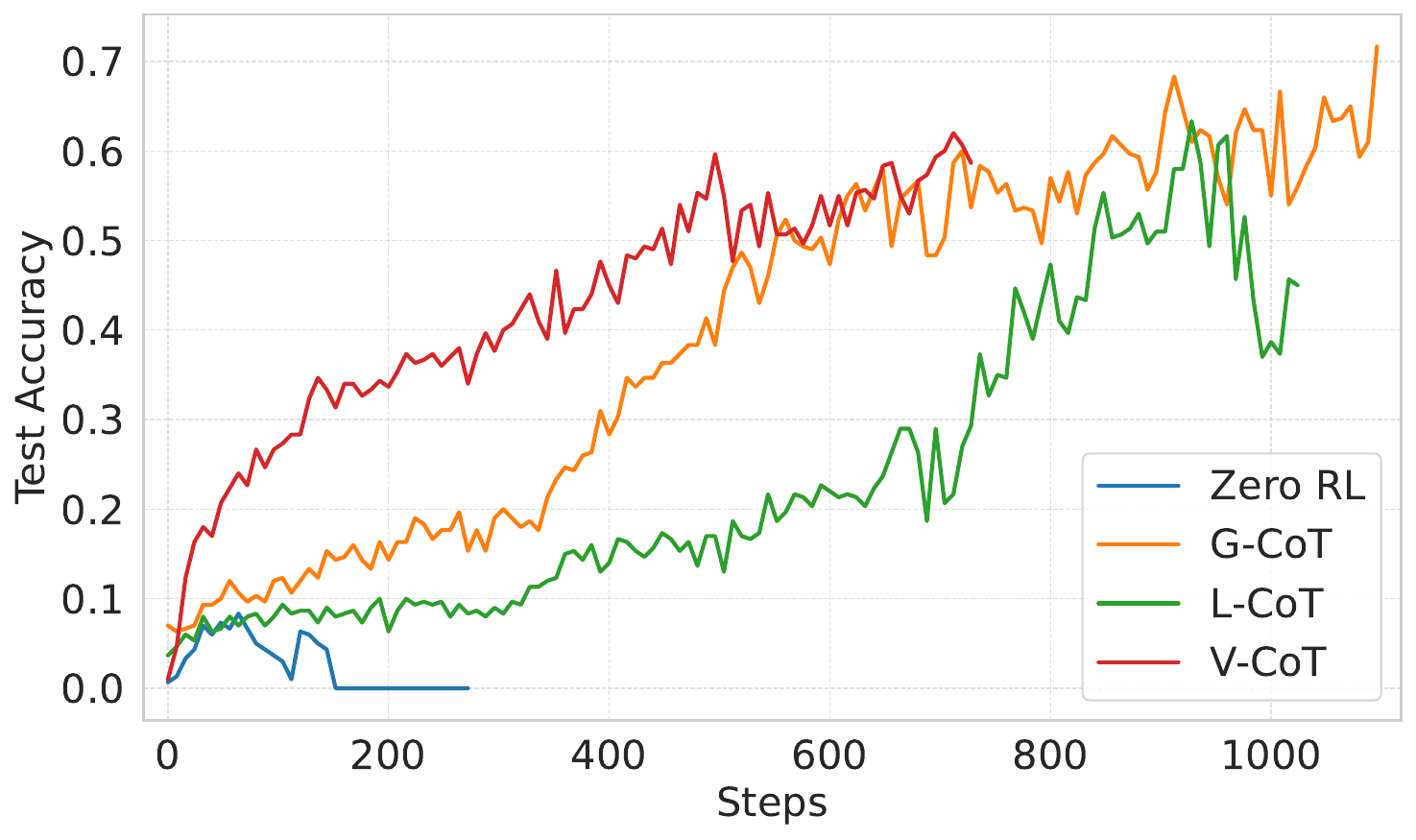}
    \caption{Test Accuracy on Mazes Sized $7\times7$.}
    
    \end{subfigure}
    \caption{Training dynamics of zero RL, Grounding CoT~(G-CoT), Languge CoT~(L-CoT), and Visual CoT~(V-CoT). The model is cold-started on mazes sized $4\times4$ to $6\times6$.}
    \label{fig:exp1}
\end{figure*}

To answer \textit{Q1}, we first compare language CoT, grounding CoT, and visual CoT under the same training setting. Specifically, we perform SFT on L-CoT, G-CoT, and V-CoT respectively to obtain three policy models, and then apply RL on the maze data. Models are trained on mazes of sized $4\times4$ to $6\times6$ and evaluated on unseen $7\times7$ mazes. The RL training dynamics in Figure~\ref{fig:exp1} lead to three observations:

$\bullet$ \textbf{Necessity of cold-start}. An SFT initialization is essential for stable and progressive RL training: models trained from scratch tend to collapse, whereas all SFT-initialized models steadily improve and eventually reach 100\% accuracy on the training mazes. This suggests that SFT provides a reasonably shaped policy space and mitigates the exploration and reward-sparsity issues that arise when starting RL from random behavior.

$\bullet$ \textbf{Training efficiency}. Among the three CoT strategies, visual CoT yields the fastest RL convergence, requiring roughly half the training steps of language CoT. A plausible explanation is that visual CoT offers more structured, spatially grounded supervision, which aligns better with the underlying maze geometry and thus provides a more informative optimization signal.

$\bullet$ \textbf{Performance upper bound}. Despite its faster convergence and higher initial performance, visual CoT does not surpass grounding CoT or language CoT in final accuracy. This indicates that visual CoT mainly accelerates optimization rather than expanding the model’s ultimate reasoning capacity: once RL has distilled an efficient internal policy for maze navigation, additional visual manipulation offers diminishing returns on asymptotic performance.

\begin{tcolorbox}[
    colframe=takeaway,
    colback=white,
    coltitle=takeawayTitle,
]
\textcolor{takeawayTitle}{\textbf{Takeaway Findings}}

Visual CoT does not offer a higher performance upper bound compared to language or grounding CoT, but only accelerates training.
\end{tcolorbox}

\subsubsection{Short CoT Surpasses Longer Ones}
\label{sec:exp2}
\begin{figure*}
    \centering
    \begin{subfigure}{0.32\linewidth}
    \includegraphics[width=1\linewidth]{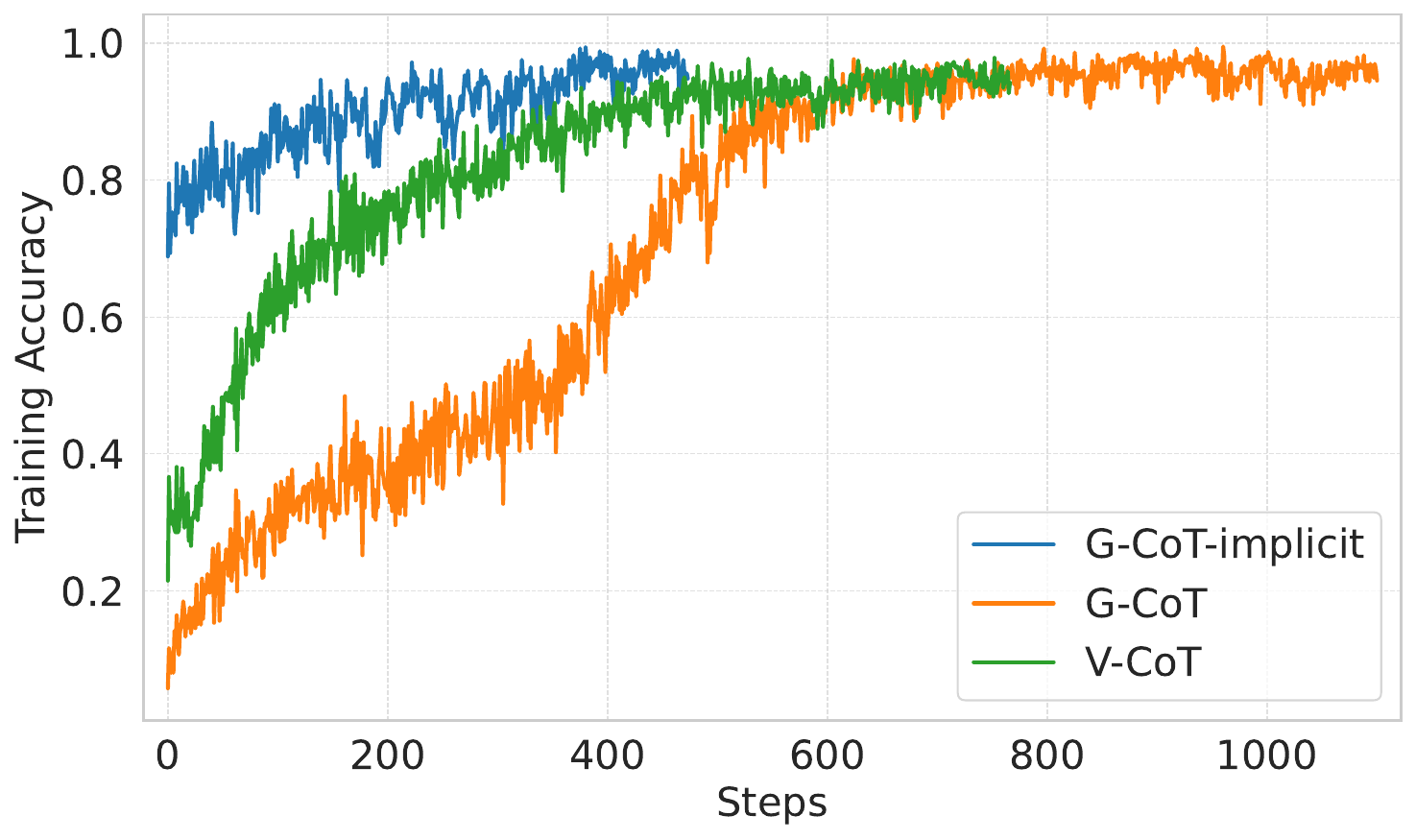}
    \caption{Training Accuracy on Mazes Sized $4\times4$ to $6\times6$.}
    
    \end{subfigure}
    \begin{subfigure}{0.32\linewidth}
    \includegraphics[width=1\linewidth]{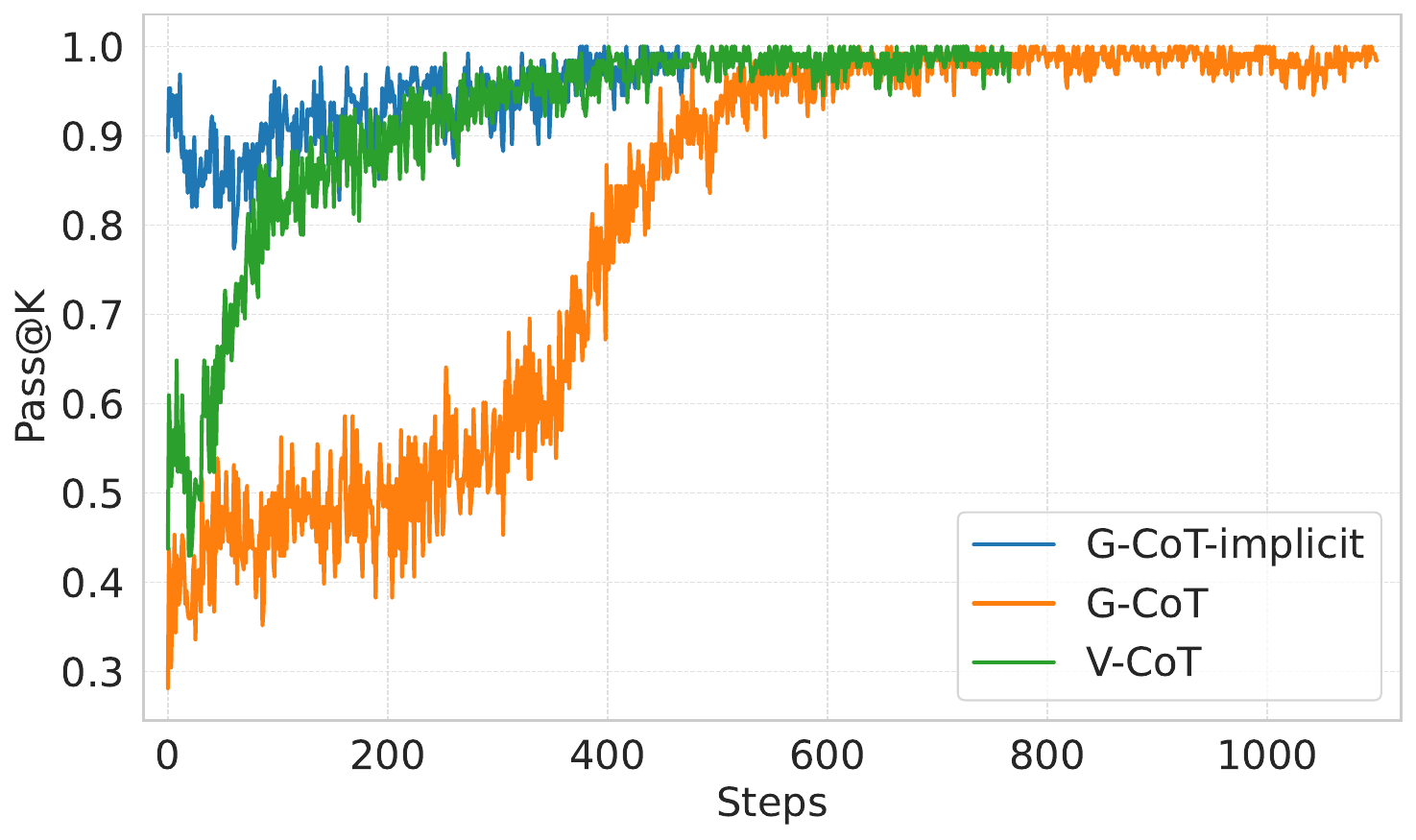}
    \caption{Pass@8 Training Accuracy on Mazes Sized $4\times4$ to $6\times6$.}
    
    \end{subfigure}
    \begin{subfigure}{0.32\linewidth}
    \centering
    \includegraphics[width=1\linewidth]{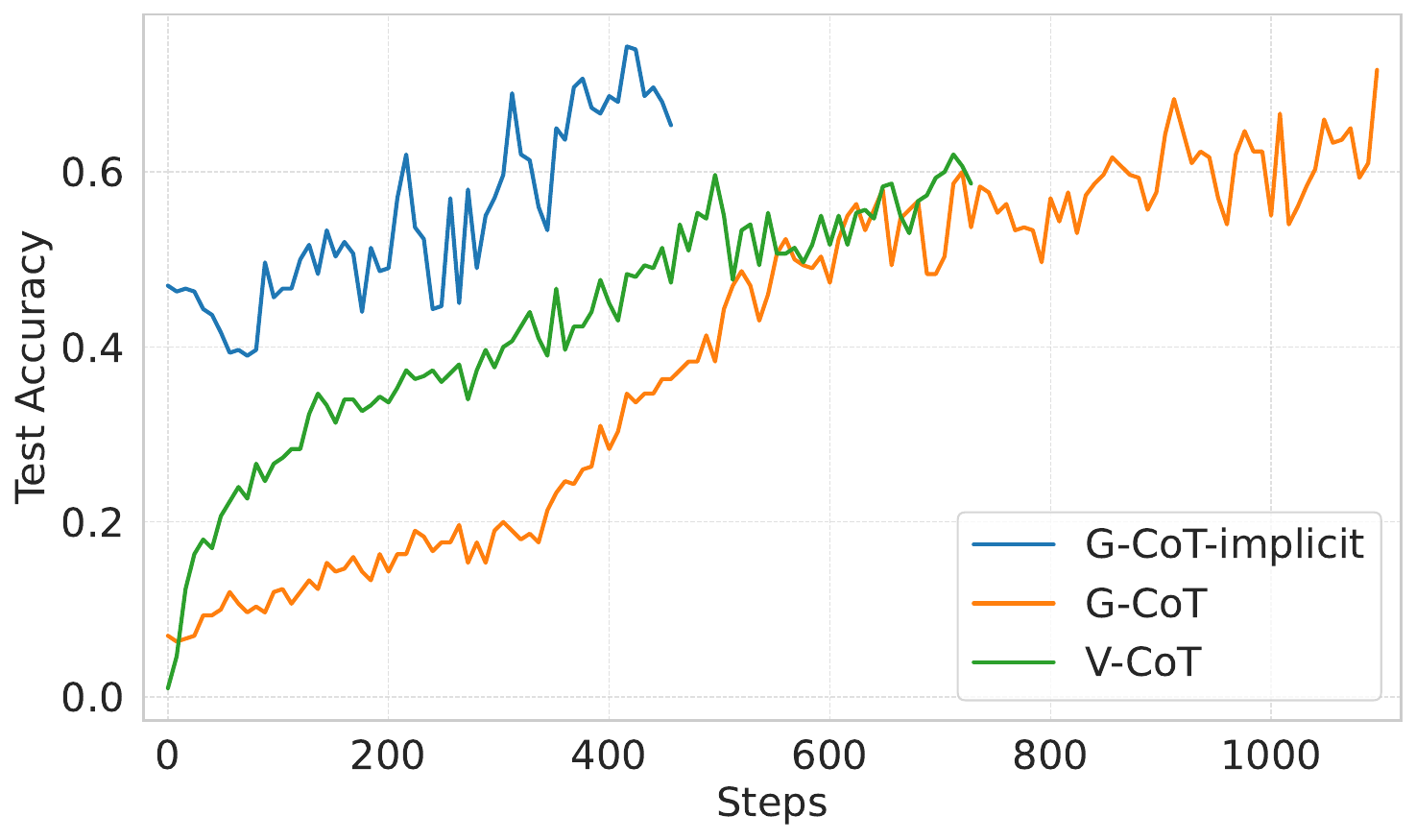}
    \caption{Test Accuracy on Mazes Sized $7\times7$.}
    
    \end{subfigure}
    \caption{Training Dynamics of least Grounding CoT~(G-CoT-least), Grounding CoT~(G-CoT), and Visual CoT~(V-CoT). The model is cold-started on mazes sized $4\times4$ to $6\times6$.}
    \label{fig:exp2}
\end{figure*}

Section~\ref{sec:exp1} shows that grounding CoT attains performance comparable to visual CoT, suggesting that the model’s reasoning is largely mediated by its grounding capability. Here, we further ask whether this grounding-based reasoning depends on the explicit coordinate system used during pre-training. To probe this, we fine-tune Qwen2.5-VL-7B on both G-CoT and G-CoT-least datasets and then apply RL under the same training and evaluation setup as in Section~\ref{sec:exp1}. As shown in Figure~\ref{fig:exp2}, the model trained with the shorter \emph{least} grounding CoT not only starts from a higher performance level but also converges faster than the one trained with explicit grounding CoT, even surpassing the efficiency of visual CoT. At the same time, it still reaches 100\% accuracy after RL, despite never seeing explicit coordinates.
These results suggest that once the model’s grounding ability is properly aligned with the visual environment, it can internalize and operate over its own latent spatial representations without relying on externally specified coordinate systems. Least grounding may thus serve as a more flexible and compact inductive bias, avoiding overfitting to a particular coordinate frame while still supporting robust, generalizable visual reasoning.


\begin{tcolorbox}[
    colframe=takeaway,
    colback=white,
    coltitle=takeawayTitle,
]
\textcolor{takeawayTitle}{\textbf{Takeaway Findings}}

The VLM can conduct implicit reasoning after its grounding ability is aligned with the visual environment, achieving full accuracy and faster convergence.
\end{tcolorbox}

\subsubsection{Least CoT Achieves Better Generalization}
\begin{figure*}
    \centering
    \begin{subfigure}{0.32\linewidth}
    \includegraphics[width=1\linewidth]{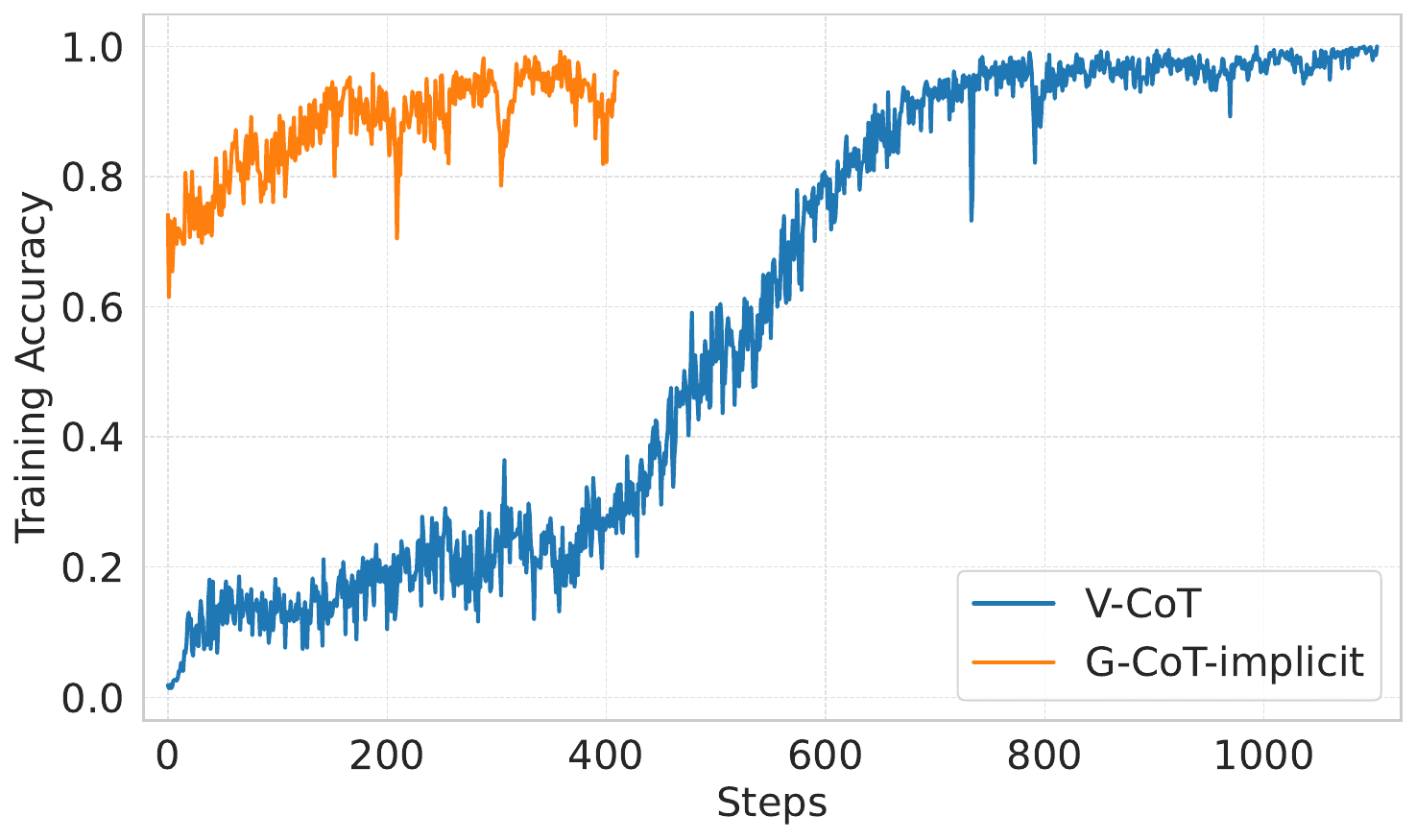}
    \caption{Training Accuracy on Mazes Sized $6\times6$.}
    
    \end{subfigure}
    \begin{subfigure}{0.32\linewidth}
    \includegraphics[width=1\linewidth]{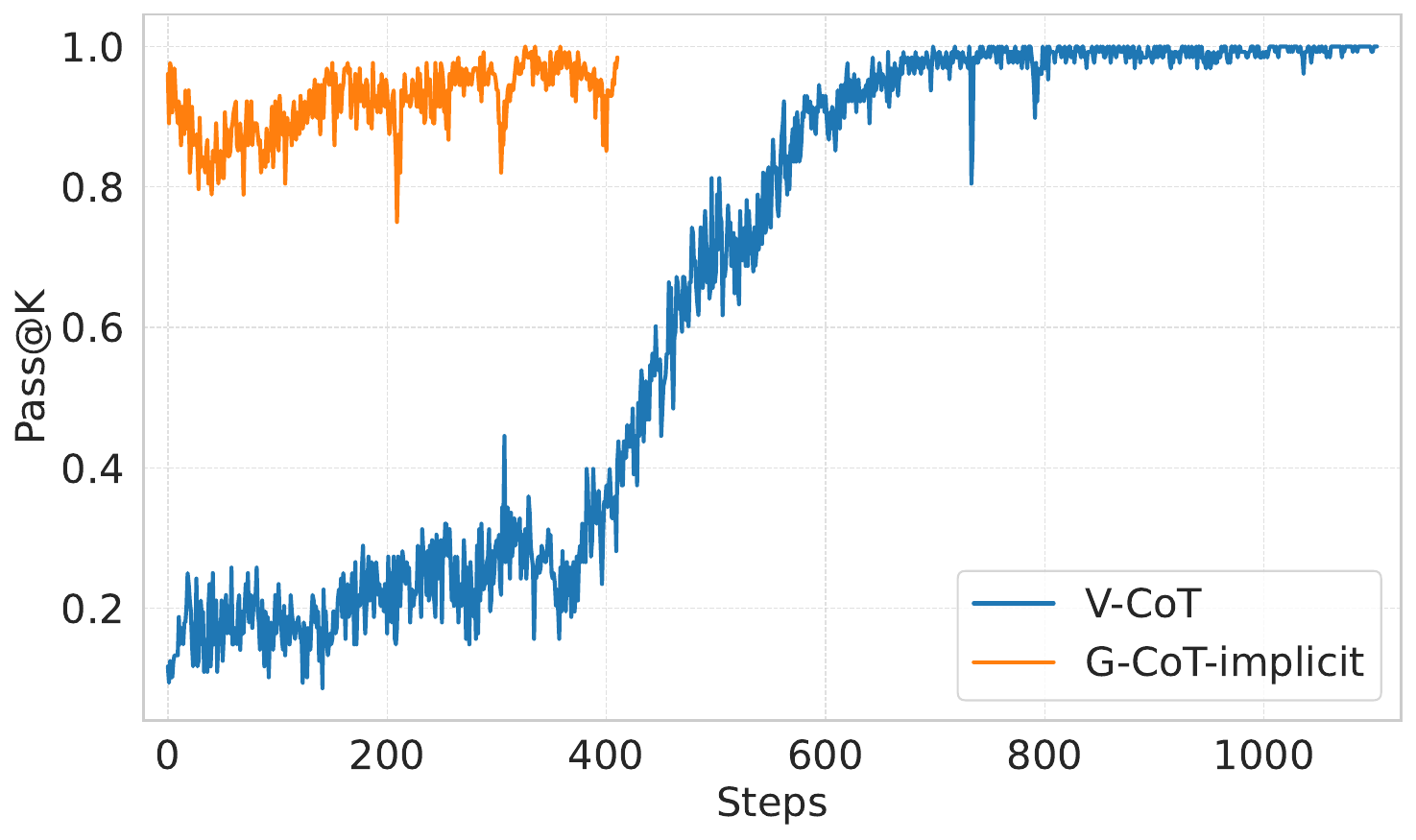}
    \caption{Pass@8 Training Accuracy on Mazes Sized $6\times6$.}
    
    \end{subfigure}
    \begin{subfigure}{0.32\linewidth}
    \centering
    \includegraphics[width=1\linewidth]{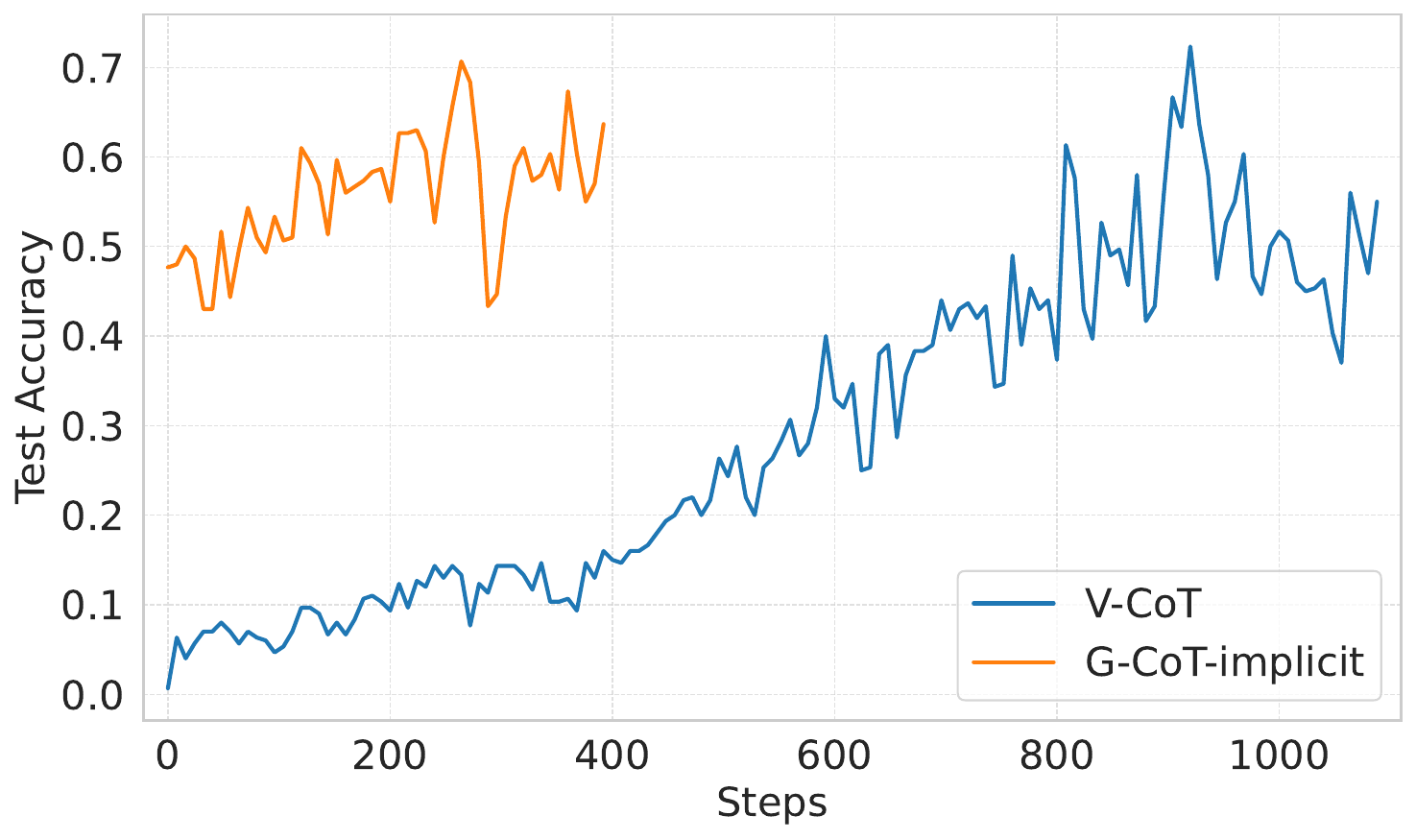}
    \caption{Test Accuracy on Mazes Sized $7\times7$.}
    
    \end{subfigure}
    \caption{Training Dynamics of least Grounding CoT~(G-CoT-least) and Visual CoT~(V-CoT). The model is cold-started on mazes sized $6\times6$ to validate the single-scale generalization.}
    \label{fig:exp3}
\end{figure*}

\begin{figure*}
    \centering
    \begin{subfigure}{0.32\linewidth}
    \includegraphics[width=1\linewidth]{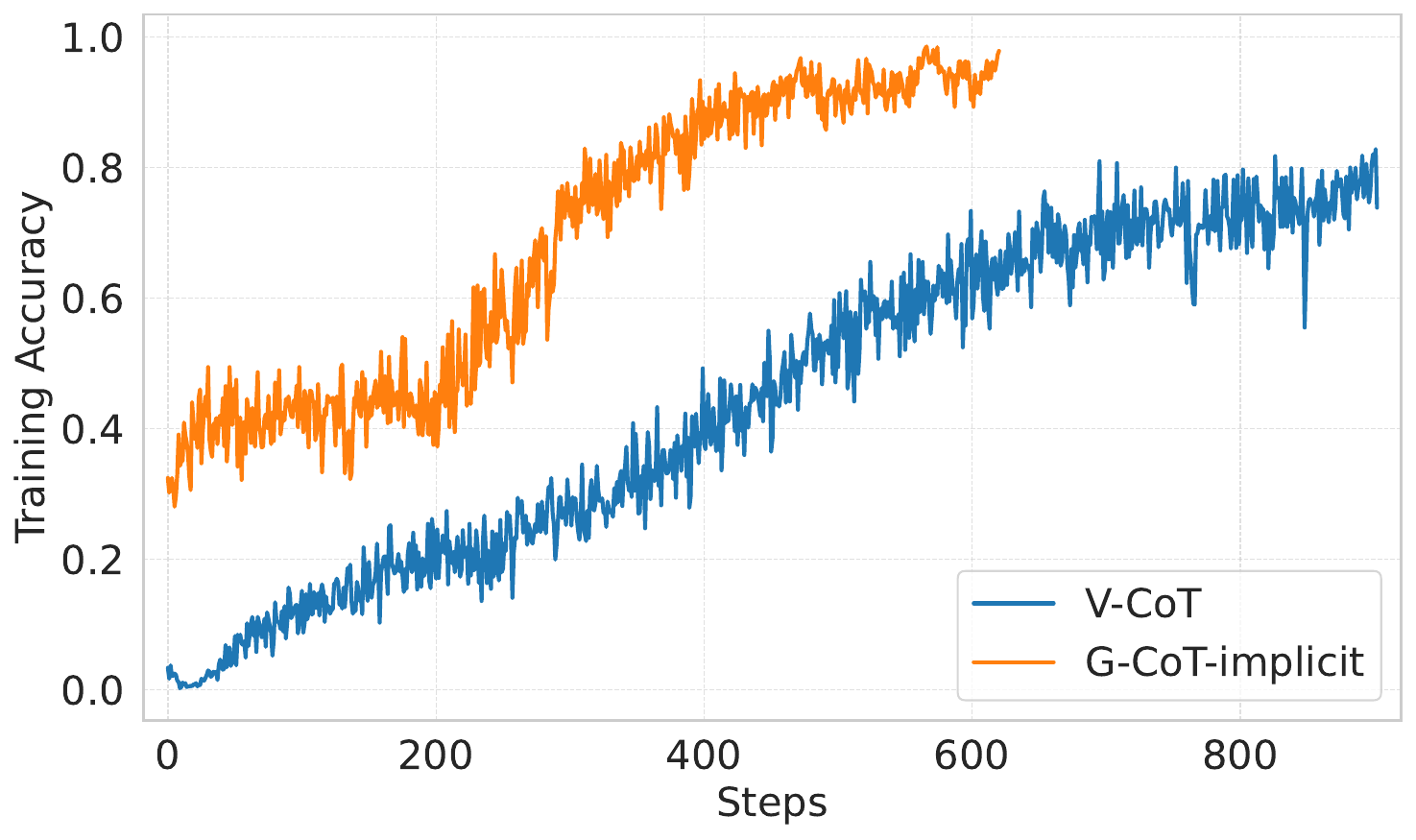}
    \caption{Training Accuracy on Mazes Sized $7\times7$ to $9\times9$.}
    
    \end{subfigure}
    \begin{subfigure}{0.32\linewidth}
    \includegraphics[width=1\linewidth]{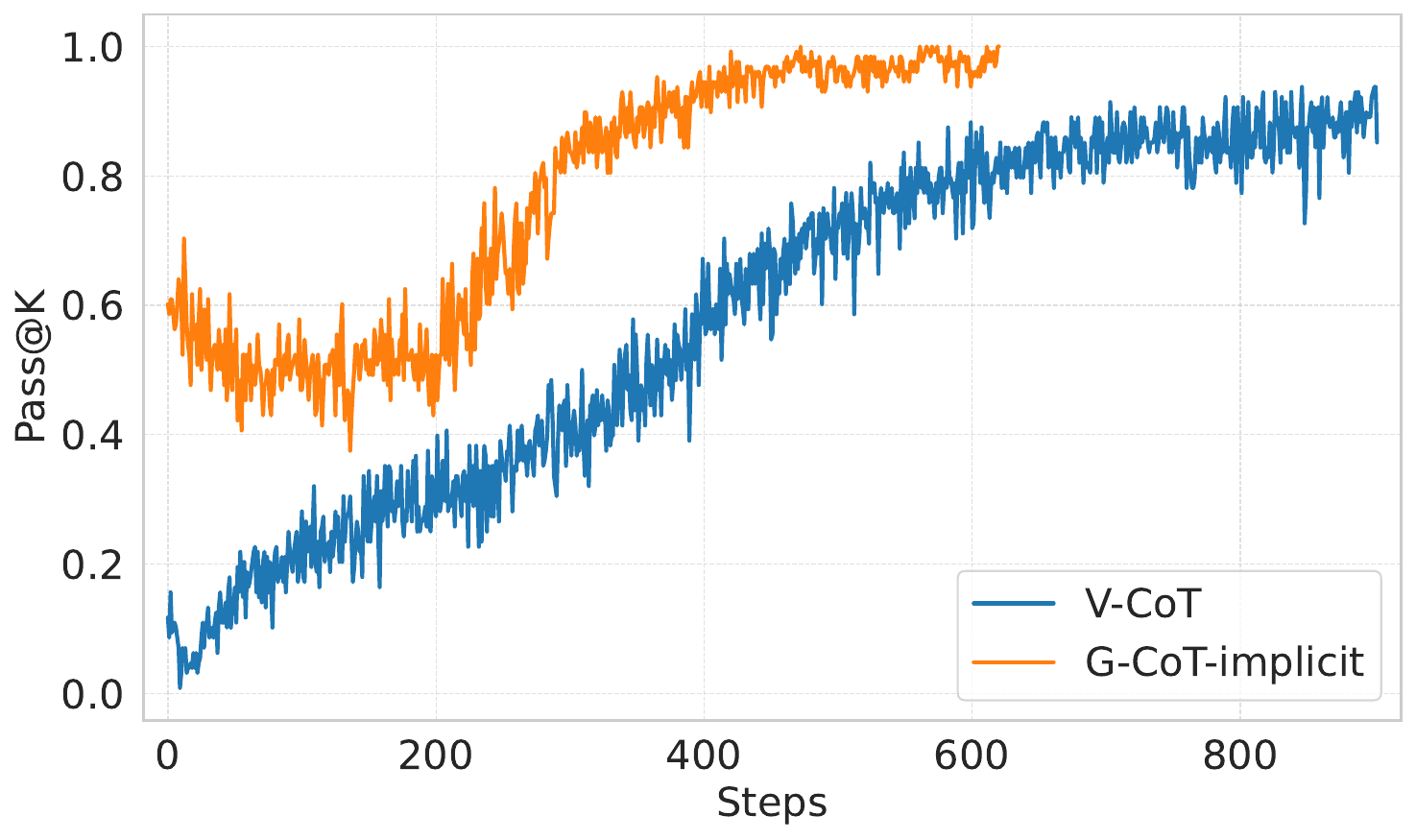}
    \caption{Pass@8 Training Accuracy on Mazes Sized $7\times7$ to $9\times9$.}
    
    \end{subfigure}
    \begin{subfigure}{0.32\linewidth}
    \centering
    \includegraphics[width=1\linewidth]{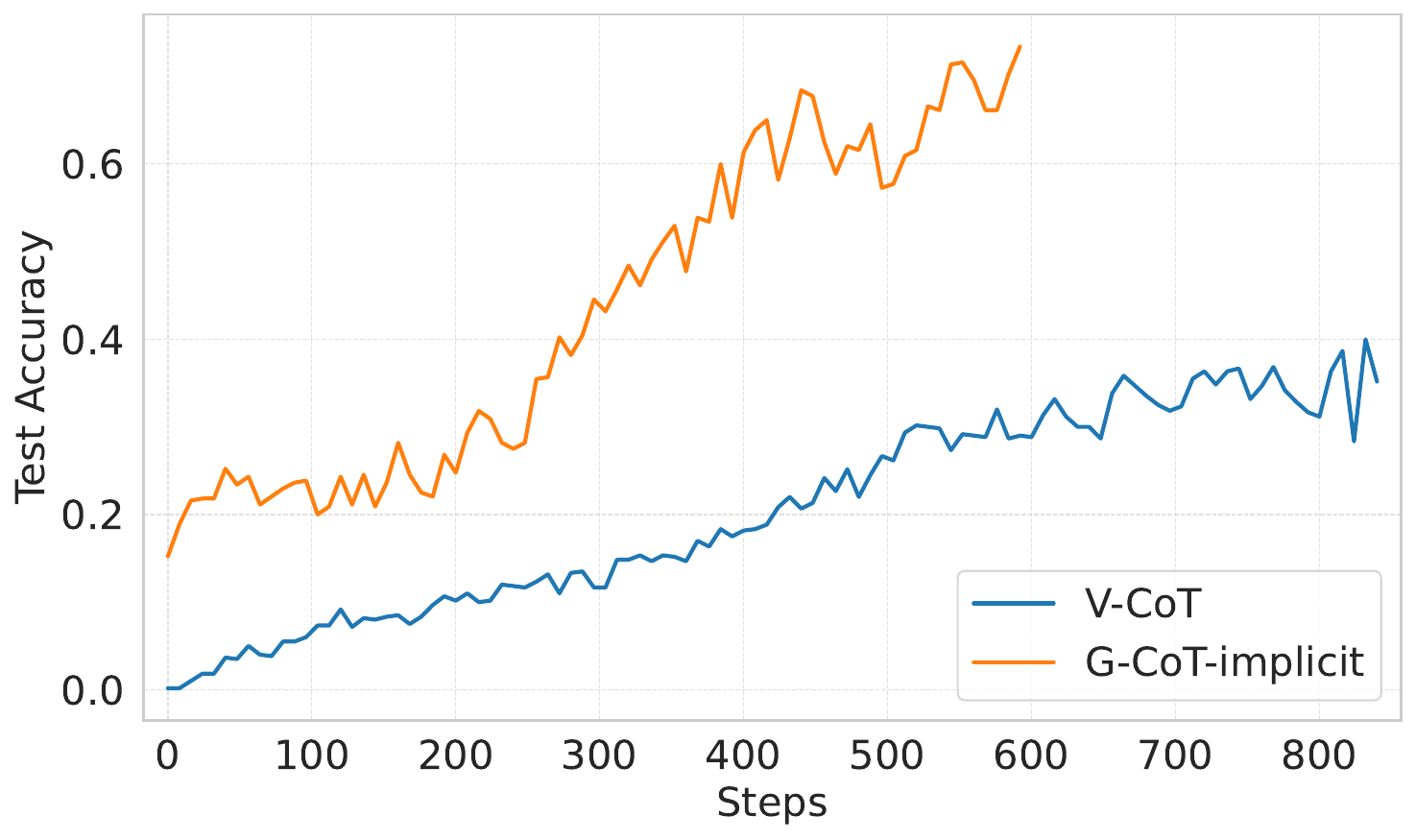}
    \caption{Test Accuracy on Mazes Sized $10\times10$.}
    
    \end{subfigure}
    \caption{Training Dynamics of least Grounding CoT~(G-CoT-least) and Visual CoT~(V-CoT). The model is cold-started on mazes sized $4\times4$ to $6\times6$ to validate the cross-scale generalization.}
    \label{fig:exp4}
\end{figure*}

\label{sec:exp3}
Our previous experiments show that the model can perform implicit reasoning through its grounding ability. We now ask whether this capability can generalize to mazes of different sizes, and evaluate two types of generalization:

$\bullet$ \textbf{Single-scale generalization:} We examine whether a model trained on mazes of a single size can generalize to slightly larger mazes. Concretely, we perform both SFT and RL on $6\times 6$ mazes and evaluate the resulting model on unseen $7\times7$ mazes.

$\bullet$ \textbf{Cross-scale generalization:} We further test whether the model can extrapolate to completely unseen maze sizes after RL, when exposed to diverse scales during SFT. Specifically, we conduct SFT on mazes sized $4\times 4$ to $6\times 6$, perform RL on mazes sized $7\times 7$ to $9\times 9$, and then evaluate on unseen $10\times 10$ mazes.

As shown in Figure~\ref{fig:exp3} and Figure~\ref{fig:exp4}, the G-CoT-least policy model generalizes robustly in both settings, maintaining high success rates on unseen maze sizes, whereas Visual CoT saturates after about 800 RL steps and remains inferior to G-CoT-least on both training and test mazes. This suggests that grounding-based implicit reasoning encourages the model to internalize scale-invariant, local navigation rules (\eg, following corridors, backtracking from dead-ends), while visual CoT is more prone to overfitting to specific visual layouts or operation patterns. By relying on compact grounded signals rather than explicit visual edits, G-CoT-least provides a more fundamental and transferable spatial representation, supporting consistent generalization to larger and more complex environments.

\begin{tcolorbox}[
    colframe=takeaway,
    colback=white,
    coltitle=takeawayTitle,
]
\textcolor{takeawayTitle}{\textbf{Takeaway Findings}}

Properly aligned grounding ability allows the model to generalize its spatial reasoning effectively to new visual environments.
\end{tcolorbox}

\subsection{Experimental Results on Other Tasks}
\label{sec:real}
\begin{table*}
    \centering
    \caption{Evaluation results on other vision reasoning tasks, where the best-performed results are marked as bond.}
    \begin{tabular}{l|ccc|ccc|c|c}
    \toprule
         \multirow{2}{*}{Model}&  \multicolumn{3}{c|}{$V^*$ Bench}&  \multicolumn{3}{c|}{HR-Bench 4K}&\multirow{2}{*}{Frozenlake} &\multirow{2}{*}{Jigsaw}\\
         &  Attr&  Spatial&  Overall&  FSP&  FCP& Overall  & &\\
    \midrule
         Qwen2.5-VL-7B&  67.83&  78.95&  72.25&  88.00&  57.00&   72.50& 20.00&0.00\\
         + V-CoT RL&  86.09&  78.95&  83.25&  87.00&  57.00&   72.00&- &-\\
         + G-CoT-least RL&  \textbf{87.83}&  \textbf{82.89}&  \textbf{85.86}&  \textbf{90.75}&  \textbf{57.50}&   \textbf{74.12}&\textbf{90.33} &\textbf{75.60}\\
    
    \bottomrule
    \end{tabular}
    
    \label{tab:real}
\end{table*}

The previous experiments on the maze dataset demonstrate that the model can conduct implicit reasoning via its grounding ability without the need to generate coordinates explicitly. In this part, we extend our analysis to more realistic vision-centric tasks to verify whether the observed mechanisms and benefits generalize beyond the maze setting. Specifically, we expand the experiments to visual games and real-world VQA tasks.

\paragraph{Visual Games.} We evaluate the effectiveness of G-CoT-least on two classic visual reasoning games: FrozenLake and Jigsaw. In FrozenLake, the model must find a path from the start point to the gift while avoiding all holes. The standard environment introduces stochasticity by allowing the agent to slip with a certain probability, but for stability and reproducibility, we set the slipping probability to 0. Since multiple optimal paths may exist, we adopt the environment implementation from~\citet{towers2024gymnasium} to reliably determine whether the agent reaches the target cell. All training and evaluation are conducted on $4\times4$ grids. In FrozenLake, G-CoT-least corresponds to a series of actions from the start point to the gift (\eg, [``Left'', ``Up'', ``Right'', ``Down'']). In jigsaw, the model must assemble nine puzzle pieces into a coherent image arranged in a $3\times3$ grid. We use the dataset provided by~\citet{wu2025visual} and follow their $3\times3$ configuration for both training and evaluation. G-CoT-least in jigsaw task corresponds to the tile indices in raster-scan order (\eg, [7, 5, 1, 6, 8, 3, 9, 2, 4]). Consistent with our maze experiments, we employ a two-stage SFT-then-RL training pipeline. The results in Table~\ref{tab:real} demonstrate that G-CoT-least significantly improves model performance on both tasks, with the gains being especially pronounced on Jigsaw, where accuracy rises from 0\% to over 70\%, demonstrating a substantial enhancement in visual reasoning capability.

\paragraph{Real-world VQA.} We follow the experimental setup of DeepEyes~\cite{zheng2025deepeyes}, performing zero-shot RL on the $V^*$ training set~\cite{tong2024eyes}. Unlike visual game environments, a model’s grounding ability is largely established during pre-training on massive collections of natural images, so additional SFT is unnecessary. We evaluate the resulting model on both the $V^*$~\cite{tong2024eyes} test set and HR-Bench~\cite{wang2025divide}. In these tasks, V-CoT refers to cropping the region of interest and appending it to the model’s context~\cite{o3,zheng2025deepeyes,zhang2025thyme}. In contrast, G-CoT-least directly answers the visual questions without introducing explicit visual CoT steps. We compare the performance of V-CoT and G-CoT-least, with results summarized in Table~\ref{tab:real}. Across all benchmarks and sub-tasks, the G-CoT-least variants achieve the best performance, demonstrating that the model can perform visual reasoning implicitly, without relying on explicit visual CoT.

\section{Related Work}

\paragraph{Vision-centric Reasoning.} Existing visual reasoning tasks can be broadly categorized into two types. The first involves tasks where reasoning can be entirely performed in the linguistic space once the visual content is understood, such as geometric problem solving~\cite{chen2021geoqa,zhang2024mathverse}, physical reasoning~\cite{chow2025physbench}, or chart understanding~\cite{masry2022chartqa,wang2024charxiv}. These are language-dominant reasoning tasks, where vision primarily serves as contextual grounding for symbolic inference. The second category comprises tasks that require active visual exploration and manipulation of the image to extract and reason over spatial clues, such as maze navigation~\cite{ivanitskiy2023configurable}, puzzle solving~\cite{xu2025visulogic}, and jigsaw~\cite{wu2025visual,wang2025jigsaw}. In these vision-centric reasoning tasks, the reasoning process is grounded in perceptual understanding rather than linguistic abstraction. This work focuses on the latter category. We use maze navigation as a controlled testbed for studying vision-centric reasoning and extend our analysis to a broader set of visual games and real-world vqa tasks.

\paragraph{Chain-of-thought Reasoning.} Chain-of-Thought (CoT) reasoning was first introduced in~\cite{wei2023cot} as an emergent capability of LLM~\cite{zhao2023survey,llama3,qwen2.5}, enabling them to decompose complex problems into intermediate reasoning steps. This approach has been shown to significantly enhance performance across a range of challenging reasoning tasks~\cite{gsm8k,jain2024livecodebench}. Recent studies on long CoT~\cite{guha2025openthoughts} further demonstrates that scaling the length of the reasoning trace enhances model performance, particularly in domains requiring multi-step inference, such as math reasoning~\cite{math} and code generation~\cite{livecodebench}. Building upon CoT for LLMs, researchers have proposed more expressive CoT formulations for VLMs, including grounding-based CoT~\cite{shao2024visual,wang2025traceable} and visual CoT~\cite{o3,zheng2025deepeyes,su2025pixel}. These variants enable models to reference specific objects or even manipulate visual content during the reasoning process. In this work, we conduct a systematic comparison of these CoT variants under a controlled experimental setting to understand their underlying mechanisms.

\paragraph{Reinforcement Learning in Vision Language Models.} Building on the success of reinforcement learning (RL) in significantly enhancing the reasoning abilities of large language models~\cite{o1,deepseekai2025deepseekr1incentivizingreasoningcapability,team2025kimi}, recent efforts have begun extending RL to VLMs~\cite{seed2025seed1_5vl,team2025kwai}. Prior work explores both zero-RL and cold-start RL regimes, showing substantial improvements in multimodal mathematical and STEM reasoning~\cite{wang2025vl,wang2025sota,deng2025openvlthinker}. For vision-centric tasks, existing studies show that RL can enhance spatial reasoning~\cite{wu2025reinforcing,sarch2025grounded}, object recognition~\cite{lithink}, and puzzle-solving abilities~\cite{feng2025visualsphinx,wang2025jigsaw}. Interestingly, these works also observe that, unlike language-dominant tasks, the CoT trajectories induced by RL in vision-centric settings tend to be short. Our work further reveals the underlying mechanism: in vision-centric environments, RL primarily strengthens the model’s established grounding capability. Once grounding is sufficiently reinforced, the model can perform effective reasoning with very brief CoT traces.

\section{Conclusion}
In this work, we systematically investigated how different Chain-of-Thought (CoT) formats affect the acquisition of \emph{generalizable} visual reasoning in vision-language models. Using a controlled maze-solving benchmark and an SFT-then-RL training paradigm on Qwen2.5-VL-7B, we compared language CoT, grounding CoT, and visual CoT. Our experiments show that visual and longer CoT traces can accelerate convergence but do not substantially raise the final performance ceiling; short CoT often surpasses longer ones; and, most notably, CoT formats containing only the least grounded information (\eg, sparse coordinate paths) generalize best across different maze sizes and related visual reasoning tasks.

These findings reveal a ``short is long'' effect: concise but well-grounded supervision appears more effective for learning reusable visual reasoning patterns than verbose, heavily supervised traces. Looking forward, we plan to extend this analysis to richer task families beyond mazes and VLMs. 

\section*{Acknowledgement}
We sincerely thank Renrui Zhang, Xiaoying Zhang, Jialong Wu, Xinchen Zhang, and other colleagues at ByteDance Seed for their support of this project.

This project is conducted solely for academic research purposes. The findings are independent of ByteDance’s commercial products and are not incorporated into any ByteDance products or commercial applications.


\bibliographystyle{plainnat}
\bibliography{main}

\clearpage

\beginappendix


\section*{Data Synthesis}
In Section~\ref{sec:exp_set}, we construct the maze dataset and synthesize three types of CoT data: language CoT, grounding CoT, and visual CoT. In this section, we introduce how we synthesize these CoT data.

\begin{figure*}
    \centering
    \includegraphics[width=1\linewidth]{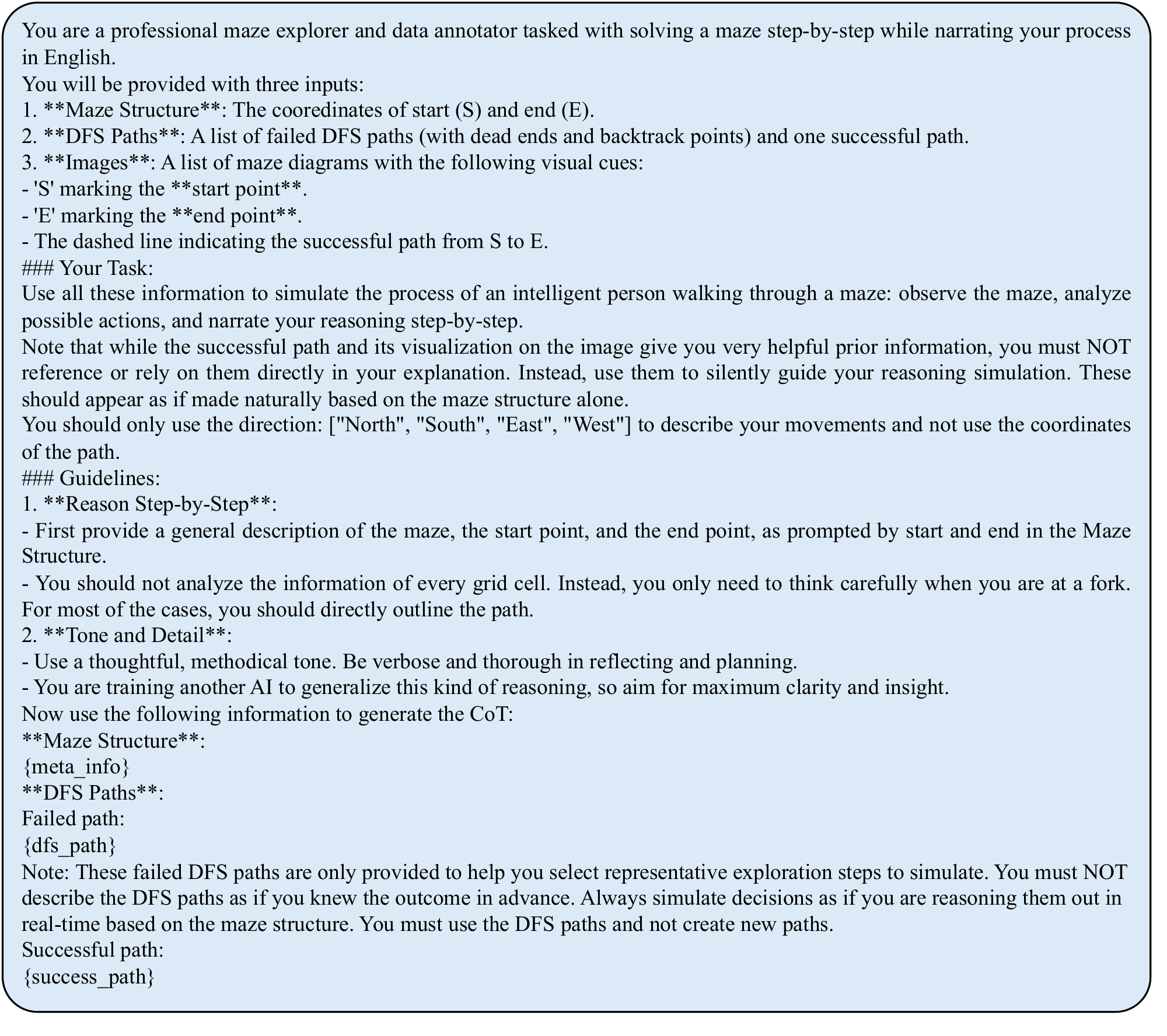}
    \caption{Prompt for language CoT synthesis.}
    \label{fig:prompt1}
\end{figure*}

\paragraph{Language CoT Synthesis.} To synthesize reasoning trajectories with language CoT, we first utilize a rule-based function to convert the path $P=[(i_s, j_s), (i_2, j_2), \dots, (i_e, j_e)]$ to a list of directions describing every step going from $s$ to $e$. Then we prompt Gemini-2.5-Pro~\cite{gemini2.5-pro} to synthesize the reasoning trajectories. The detailed prompt is shown in Figure~\ref{fig:prompt1}.

\begin{figure*}
    \centering
    \includegraphics[width=1\linewidth]{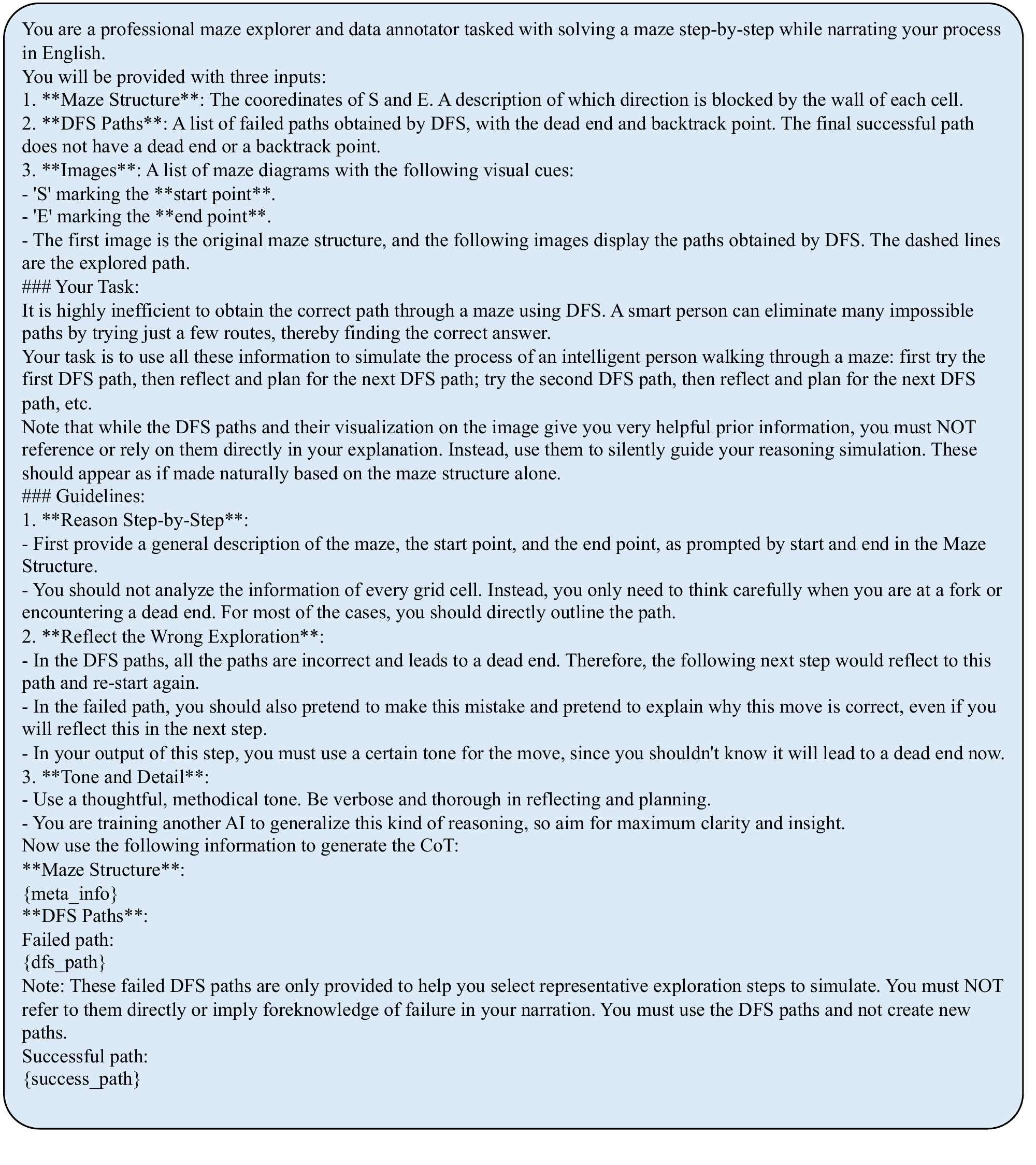}
    \caption{Prompt for grounding CoT synthesis.}
    \label{fig:prompt2}
\end{figure*}

\paragraph{Grounding CoT Synthesis.} We first utilize a rule-based function to convert each grid $(i_k, j_k)$ in the path $P=[(i_s, j_s), (i_2, j_2), \dots, (i_e, j_e)]$ to its absolute coordinate in the image $[x_k, y_k]$. Then we prompt Gemini-2.5-Pro~\cite{gemini2.5-pro} to synthesize the reasoning trajectories. The detailed prompt is shown in Figure~\ref{fig:prompt2}.

\begin{figure*}
    \centering
    \includegraphics[width=1\linewidth]{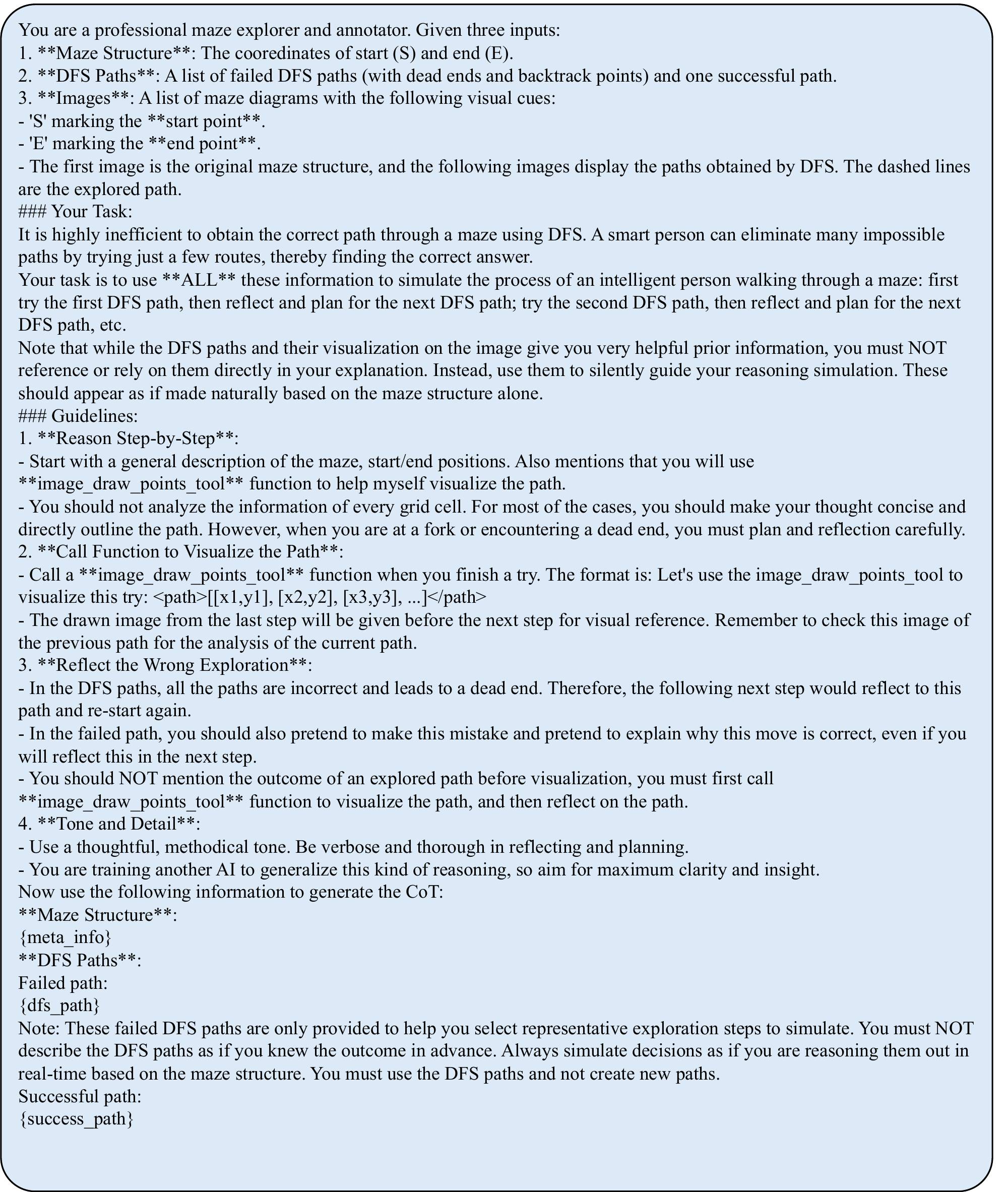}
    \caption{Prompt for visual CoT synthesis.}
    \label{fig:prompt3}
\end{figure*}

\paragraph{Visual CoT Synthesis.} To obtain coherent reflective patterns in visual CoT, we define a line-drawing function on the image and plot each point $[x_k, y_k]$ on the image. Then we prompt Gemini-2.5-Pro~\cite{gemini2.5-pro} to synthesize the reasoning trajectories. The prompt is shown in Figure~\ref{fig:prompt3}.

\end{document}